\documentclass{article}

\usepackage{microtype}
\usepackage{graphicx}
\usepackage{subfigure}
\usepackage{booktabs} %
\usepackage{amsmath}
\usepackage{duckuments}
\usepackage{mathtools}
\usepackage{makecell}
\usepackage{multirow}
\usepackage{comment}
\usepackage{graphbox}
\usepackage{tabularx}
\usepackage{float}
\usepackage{amsfonts}
\usepackage{pgfplots,pgfplotstable}
\usepgfplotslibrary{fillbetween}

\pgfplotsset{compat=1.14}
\pgfplotsset{compat/show suggested version=false}

\usepackage[normalem]{ulem}
\def\clap#1{\hbox to 0pt{\hss #1\hss}}%
\usepackage{color}
\definecolor{lightgreen}{rgb}{0, 0.6, 0.3}
\definecolor{olive}{rgb}{0.5, 0.5, 0.0}
\definecolor{maroon}{rgb}{0.69, 0.19, 0.38}
\definecolor{celestialblue}{rgb}{0.29, 0.59, 0.82}
\definecolor{darkgreen}{rgb}{0.0, 0.5, 0.0}
\definecolor{darkblue}{rgb}{0.19, 0.19, 0.62}
\definecolor{grey}{rgb}{0.5,0.5,0.5}

\newif\ifgithub
\newif\ifacknowledgments

\newcolumntype{Y}{>{\centering\arraybackslash}X}

\usepackage[unanon]{icml2021}\githubtrue\acknowledgmentstrue %

\usepackage[pagebackref=true,breaklinks=true,colorlinks,bookmarks=true, citecolor=violet]{hyperref}

\icmltitlerunning{StyleGAN-T: Unlocking the Power of GANs for Fast Large-Scale Text-to-Image Synthesis}

\newcommand{\bc}{\mathbf{c}}

\newcommand{\bs}{\mathbf{s}}

\newcommand{\bw}{\mathbf{w}}

\newcommand{\bz}{\mathbf{z}}

\newcommand{\figref}[1]{Fig.~\ref{#1}}
\newcommand{\secref}[1]{Section~\ref{#1}}
\newcommand{\shortsecref}[1]{Sec.~\ref{#1}}

\newcommand{\shorteqnref}[1]{Eq.~\ref{#1}}
\newcommand{\tabref}[1]{Table~\ref{#1}}

\makeatletter
\DeclareRobustCommand\onedot{\futurelet\@let@token\@onedot}
\def\@onedot{\ifx\@let@token.\else.\null\fi\xspace}

\makeatother

\newcommand{\boldparagraph}[1]{\vspace{0.2cm}\noindent{\bf #1} }

\definecolor{darkgreen}{rgb}{0,0.7,0}
\definecolor{darkblue}{RGB}{31,119,180}
\definecolor{darkred}{RGB}{214,39,40}

\usepackage{appendix}

\pgfplotsset{xtick style={draw=none}}
\pgfplotsset{ytick style={draw=none}}
\pgfplotsset{major grid style={gray!40}}

\definecolor{C0}{rgb}{0.121569, 0.466667, 0.705882}
\definecolor{C1}{rgb}{1.000000, 0.498039, 0.054902}
\definecolor{C2}{rgb}{0.172549, 0.627451, 0.172549}
\definecolor{C3}{rgb}{0.839216, 0.152941, 0.156863}
\definecolor{C4}{rgb}{0.580392, 0.403922, 0.741176}
\definecolor{C5}{rgb}{0.549020, 0.337255, 0.294118}
\definecolor{C6}{rgb}{0.890196, 0.466667, 0.760784}
\definecolor{C7}{rgb}{0.498039, 0.498039, 0.498039}
\definecolor{C8}{rgb}{0.737255, 0.741176, 0.133333}
\definecolor{C9}{rgb}{0.090196, 0.745098, 0.811765}

\newcommand{\teaserGroup}[2]{ %
  \addlegendimage{#1, only marks, mark size=2pt};
  \addlegendentry[#1]{\hspace{0.5mm}#2};
}

\newcommand{\teaserPoint}[5]{ %
  \addplot[#1, only marks, mark size=2pt, forget plot] coordinates {(#2,#3)};
  \node at (axis cs:#2,#3) [#1, anchor={#5}] {#4};
}

\newcommand{\teaserlog}{
\begin{figure}[t]
\centering%
\resizebox{\linewidth}{!}{%
\begin{tikzpicture}%
\begin{axis}[
  width=100mm, height=66mm,
  xlabel={Time to generate one image [s]}, xmin={0}, xmax={35}, x coord trafo/.code=\pgfmathparse{##1^0.5}, xtick={0, 1, 3, 5, 10, 15, 20, 25, 30, 35}, xticklabels={$0$, $1$, $3$, $5$, $10$, $15$, $20$, $25$, $30$, $35$},
  ylabel={Zero-shot FID\textsubscript{30k}}, ymin={5}, ymax={28}, ymode={linear}, ytick={5, 10, 15, 20, 25},
  grid={major}, legend pos={north east}, legend cell align={left},
]
\teaserGroup{C0}{Diffusion}
\teaserPoint{C0}{3.7}{12.63}{LDM}{west}
\teaserPoint{C0}{15.0}{12.24}{GLIDE}{south}
\teaserPoint{C0}{3.7}{8.59}{Stable Diffusion}{east}
\teaserPoint{C0}{9.1}{7.27}{Imagen}{west}
\teaserPoint{C0}{32.0}{6.95}{eDiff-I}{east}
\teaserGroup{C1}{Autoregressive}
\teaserPoint{C1}{25}{11.84}{Make-a-Scene}{north}
\teaserPoint{C1}{6.4}{8.1}{Parti-3B}{south}
\teaserGroup{C2}{GAN}
\teaserPoint{C2}{0.02}{26.94}{LAFITE}{west}
\teaserPoint{C2}{0.10}{13.9}{\textbf{StyleGAN-T}}{west}
\end{axis}
\end{tikzpicture}%
}%
\vspace*{-2mm}%
\caption{%
\textbf{Quality vs. speed} in large-scale text-to-image synthesis.  StyleGAN-T greatly narrows the quality gap between GANs and other model families while generating samples at a rate of 10 FPS on an NVIDIA A100.
  The $y$-axis corresponds to zero-shot FID on MS COCO at 256$\times$256 resolution; lower is better.
}%
\label{fig:teaser}
\end{figure}
}

\newcommand{\bigsamples}{
\begin{figure*}[p]
\centering%
\includegraphics[width=0.907\linewidth]{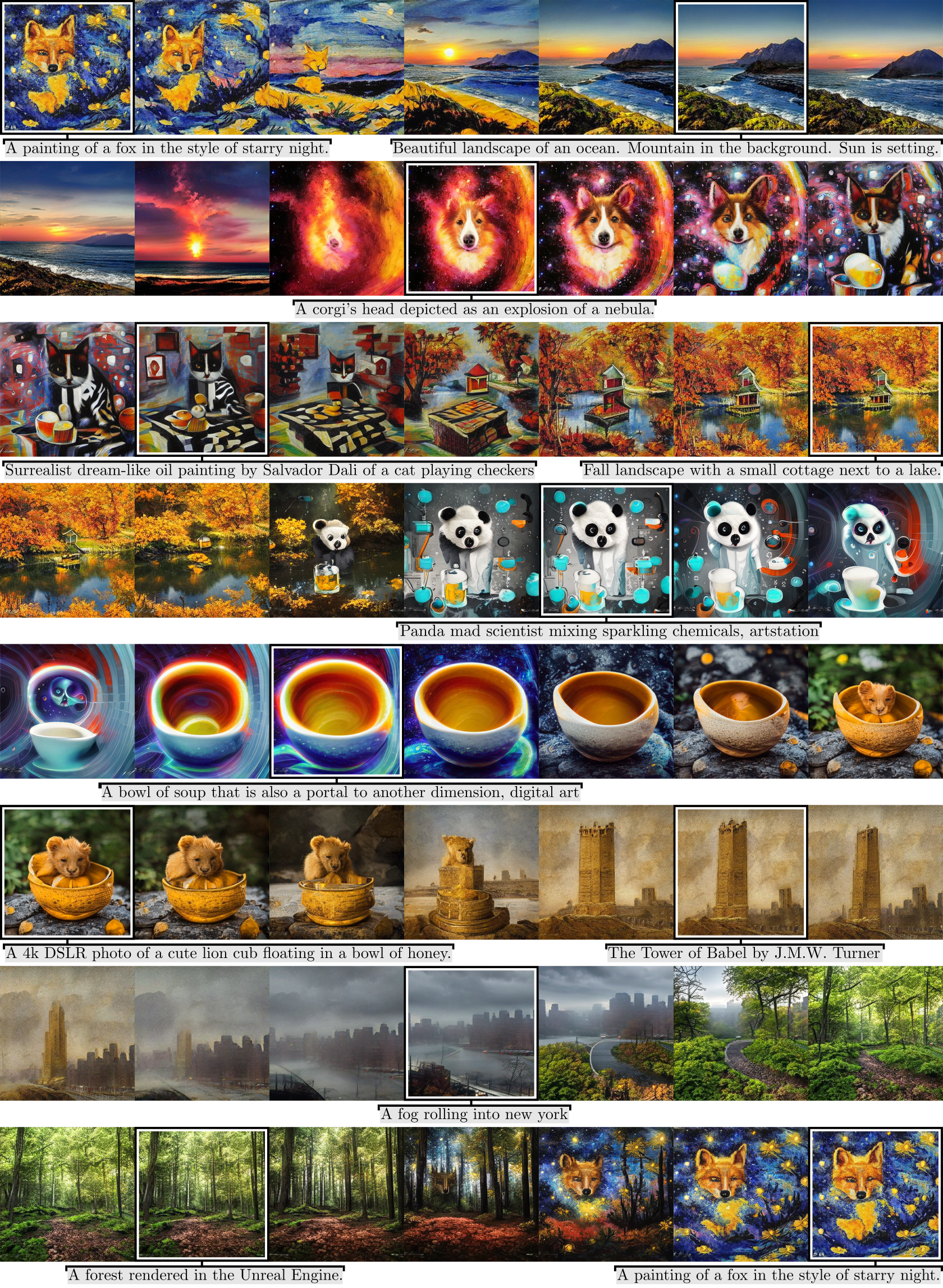}%
\vspace*{-3.5mm}%
\caption{
  \textbf{Example images and interpolations.} StyleGAN-T generates diverse samples matching the text prompt and allows for smooth interpolations between prompts, illustrated as a single continuous interpolation in scanline order. 
  Generating these 56 samples at 512$\times$512 takes~6 seconds on an NVIDIA RTX 3090, while a comparable grid takes up to several minutes with current diffusion models.
  The accompanying video further demonstrates interpolations and contrasts them with diffusion models.
}
\label{fig:bigsamples}
\end{figure*}
}

\newcommand{\system}{
\begin{figure*}[t]
\centering%
\includegraphics[width=\linewidth,trim={128 48 128 50},clip]{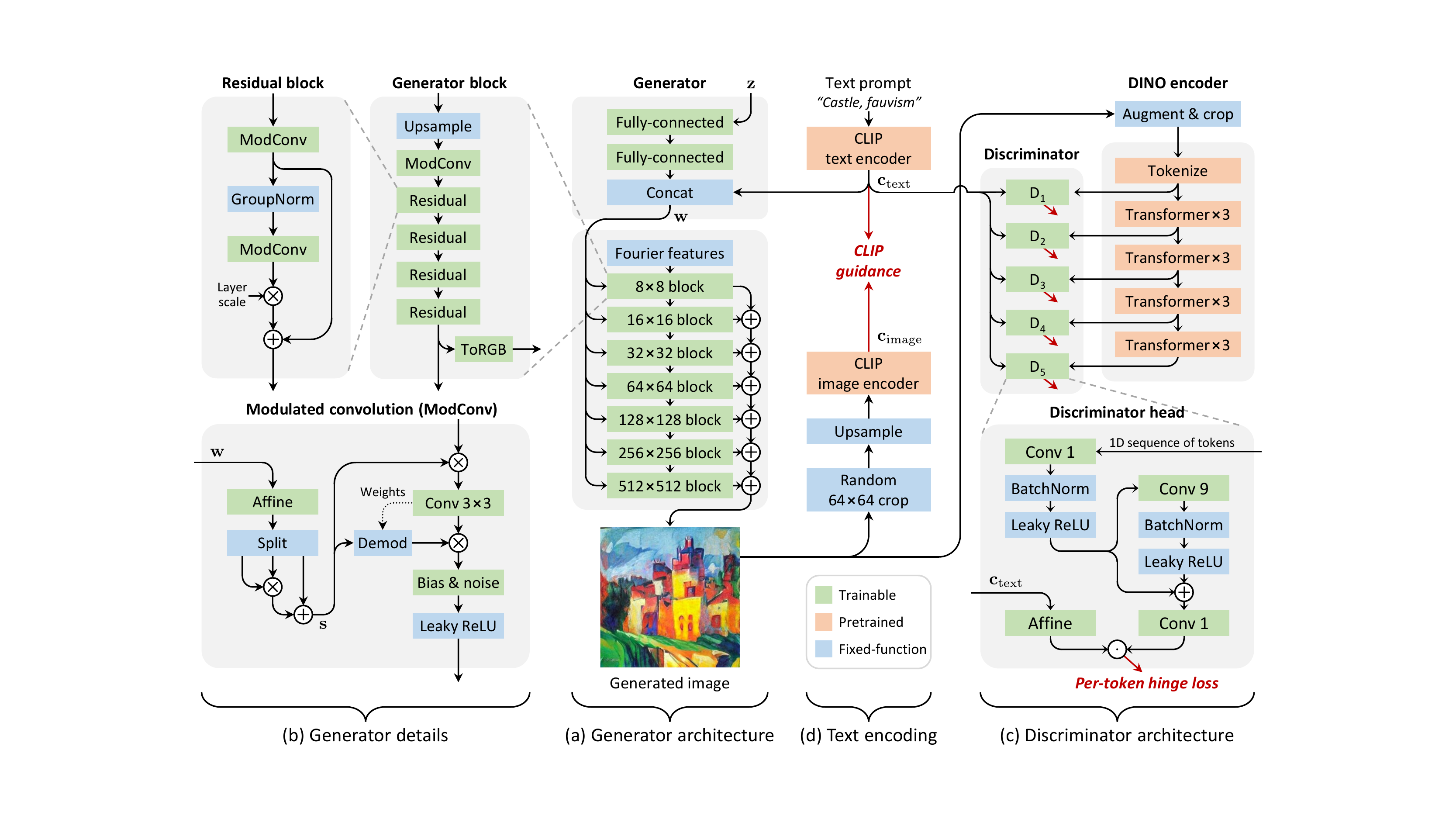}%
\vspace*{-2mm}\\%
\caption{ 
  \textbf{Overview of StyleGAN-T.}
  \textbf{(a)} Our generator architecture (\shortsecref{sec:generator}) is closely related to StyleGAN2, with the learned constant replaced with Fourier features and conditioning applied in a slightly different place.
  \textbf{(b)} For each resolution, a generator block is executed and its contribution is accumulated to the image via a dedicated ToRGB layer. The generator blocks employ residual connections and a new 2\textsuperscript{nd} order style mechanism (\shorteqnref{eq:poly}).
  \textbf{(c)} Our discriminator (\shortsecref{sec:discriminator}) processes the intermediate tokens of a DINO-trained vision transformer using 5 identical discriminator heads. Text conditioning is done using projection at the end.
  \textbf{(d)} Text prompt is embedded using CLIP and supplied to the generator and discriminator. We also employ a guidance term to further improve text alignment (\shortsecref{sec:guidance}).
}
\label{fig:system}
\end{figure*}
}

\newcommand{\tetrainingTrained}{
(0.304709415435791, 24.232327533869036)
(0.30354503631591795, 23.214245802031257)
(0.3020538139343262, 22.26808578751586)
(0.30178037643432615, 21.964200338343055)
(0.2987831497192383, 20.980434378225244)
(0.2983003044128418, 20.573519219463595)
(0.2971324348449707, 20.13091776414437)
(0.2945480537414551, 19.795256875788045)
(0.2928913116455078, 19.351840066034253)
}
\newcommand{\tetrainingFrozen}{
(0.29796194076538085, 23.63211605085263)
(0.29663156509399413, 22.21749451879616)
(0.2959703826904297, 21.5981651233384)
(0.2941876792907715, 21.098818429270494)
(0.2923, 19.9)
(0.29187744140625, 19.718952227868137)
(0.28866334915161135, 19.606352738190754)
(0.2855, 19.5)
(0.283, 19.3)
}

\newcommand{\tetrainingCurve}[3]{ %
  \addplot[#1, very thick, mark=*, mark size=1.2pt] coordinates {#2};
  \addlegendentry{\hspace{0.5mm}#3};
}

\newcommand{\tetraining}{
\begin{figure}[t]
\centering%
\resizebox{\linewidth}{!}{%
\begin{tikzpicture}%
\begin{axis}[
  width=100mm, height=55mm, %
  xlabel={CLIP score (ViT-g-14)}, xmin={0.283}, xmax={0.306}, xmode={linear}, xtick={0.285, 0.290, 0.295, 0.300, 0.305}, xticklabels={$0.285$, $0.290$, $0.295$, $0.300$, $0.305$},
  ylabel={Zero-shot FID\textsubscript{5k}}, ymin={19}, ymax={24.5}, ymode={linear}, ytick={20, 22, 24},
  grid={major}, legend pos={north west}, legend cell align={left},
]
\tetrainingCurve{C0}{\tetrainingFrozen}{TE frozen}
\tetrainingCurve{C2}{\tetrainingTrained}{TE trained}
\end{axis}
\end{tikzpicture}%
}%
\vspace*{-2mm}%
\caption{ 
  \textbf{Text encoder training.}
  Training the CLIP text encoder (TE) pushes the entire FID--CLIP score curve to the right, hence, increasing overall text alignment.
}
\label{fig:tetraining}
\end{figure}
}

\newcommand{\truncationstylegant}{
(0.304709415435791, 24.232327533869036)
(0.30354503631591795, 23.214245802031257)
(0.3020538139343262, 22.26808578751586)
(0.30178037643432615, 21.964200338343055)
(0.2987831497192383, 20.980434378225244)
(0.2983003044128418, 20.573519219463595)
(0.2971324348449707, 20.13091776414437)
(0.2945480537414551, 19.795256875788045)
(0.2928913116455078, 19.351840066034253)
}
\newcommand{\truncationediffi}{
(0.2705, 20.15)
(0.28550000000000003, 18.65)
(0.29350000000000004, 18.64)
(0.2985, 19.2)
(0.30219999999999997, 19.9)
(0.304, 20.7)
(0.306, 21.25)
(0.307, 21.9)
(0.308, 22.4)
(0.312, 26.2)
}

\newcommand{\truncationsddistilled}{
(0.287, 26.671)
(0.296, 22.162)
(0.297, 22.031)
}

\newcommand{\truncationCurve}[3]{ %
  \addplot[#1, very thick, mark=*, mark size=1.2pt] coordinates {#2};
  \addlegendentry{\hspace{0.5mm}#3};
}

\newcommand{\truncation}{
\begin{figure}[t]
\centering%
\resizebox{\linewidth}{!}{%
\begin{tikzpicture}%
\begin{axis}[
  width=100mm, height=55mm, %
  xlabel={CLIP score (ViT-g-14)}, xmin={0.26}, xmax={0.315}, xmode={linear}, xtick={0.27, 0.28, 0.29, 0.30, 0.31}, xticklabels={$0.27$, $0.28$, $0.29$, $0.30$, $0.31$},
  ylabel={Zero-shot FID\textsubscript{5k}}, ymin={18}, ymax={28}, ymode={linear}, ytick={18, 20, 22, 24, 26, 28},
  grid={major}, legend pos={north west}, legend cell align={left},
]
\truncationCurve{C0}{\truncationsddistilled}{SD-distilled}
\truncationCurve{C1}{\truncationediffi}{eDiff-I}
\truncationCurve{C2}{\truncationstylegant}{StyleGAN-T}
\end{axis}
\end{tikzpicture}%
}%
\vspace*{-2mm}%
\caption{ 
  \textbf{Comparing text alignment tradeoffs.} We compare FID--CLIP score curves of StyleGAN-T, distilled Stable Diffusion (SD-distilled), and eDiff-I.
  We report values of SD-distilled at a guidance scale of $w=4$.
  For a fair comparison, we report numbers for CLIP-conditioned eDiff-I 
  disabling additional conditioning on T5-XXL text embeddings.
  The models use different methods to increase the CLIP score (i.e., text alignment): 
  StyleGAN-T decreases truncation $\psi=\{1.0\dots0.0\}$,
  SD-distilled increases the number of sampling steps $\{2, 4, 8\}$,
  eDiff-I increases guidance scale $w=\{0\dots10\}$.
}
\label{fig:truncation}
\end{figure}
}

\newcommand{\stylesamples}{
\begin{figure}[t]
\centering%
\setlength{\tabcolsep}{0.2pt}
\begin{tabular}{ccc}
\vspace*{-1mm}%
\includegraphics[width=.33\linewidth]{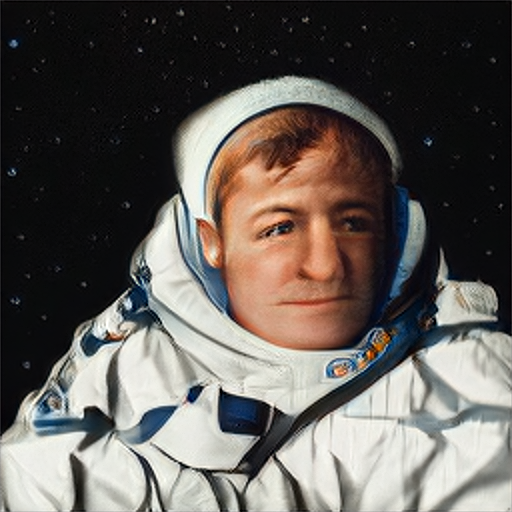} & \includegraphics[width=.33\linewidth]{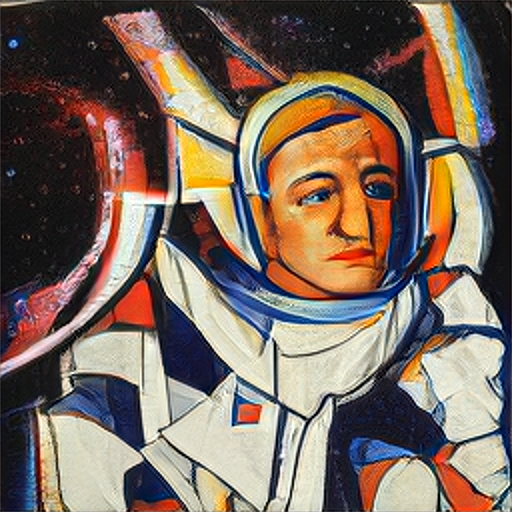} & \includegraphics[width=.33\linewidth]{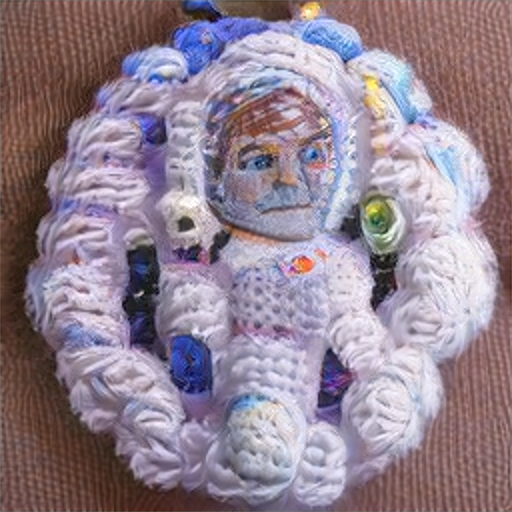} \\
\scriptsize \makecell{``real photo''} &
\scriptsize \makecell{``cubism painting''} &
\scriptsize \makecell{``made of beads and yarn''}\\
\vspace*{-1mm}
\includegraphics[width=.33\linewidth]{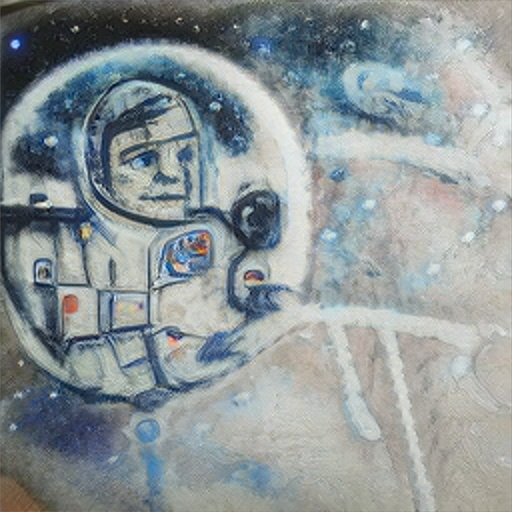} & \includegraphics[width=.33\linewidth]{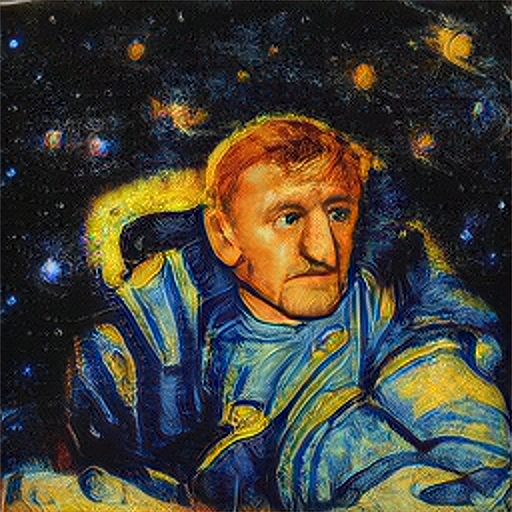} & \includegraphics[width=.33\linewidth]{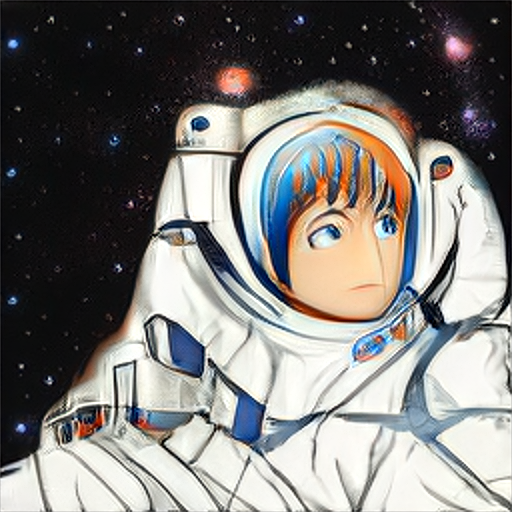} \\
\scriptsize \makecell{``chalk art''} &
\scriptsize \makecell{``van Gogh painting''} &
\scriptsize \makecell{``anime''}
\end{tabular}
\vspace*{-4mm}%
\caption{
\textbf{Styles.}
Samples generated by StyleGAN-T for a fixed random seed and the caption “astronaut, \{X\}”, where X is denoted below each image.
}%
\vspace*{-2mm}%
\label{fig:stylesamples}
\end{figure}
}

\newcommand{\truncationsamples}{
\begin{figure}[t]
    \centering
    \setlength{\tabcolsep}{0.2pt}
    \begin{tabular}{lc}
        \scriptsize \makecell{\quad$\psi = 1.00 $\\\\$\overline{CS} = 0.33$}\;&
        \includegraphics[align=c, width=.83\linewidth]{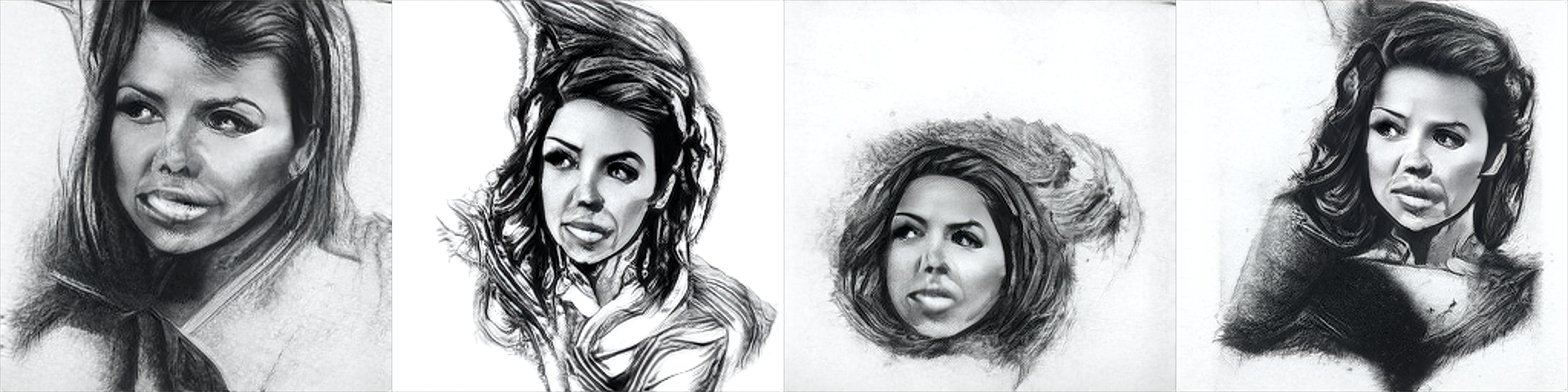}\\
        \scriptsize \makecell{\quad$\psi = 0.60 $\\\\$\overline{CS} = 0.36$}\;&
        \includegraphics[align=c, width=.83\linewidth]{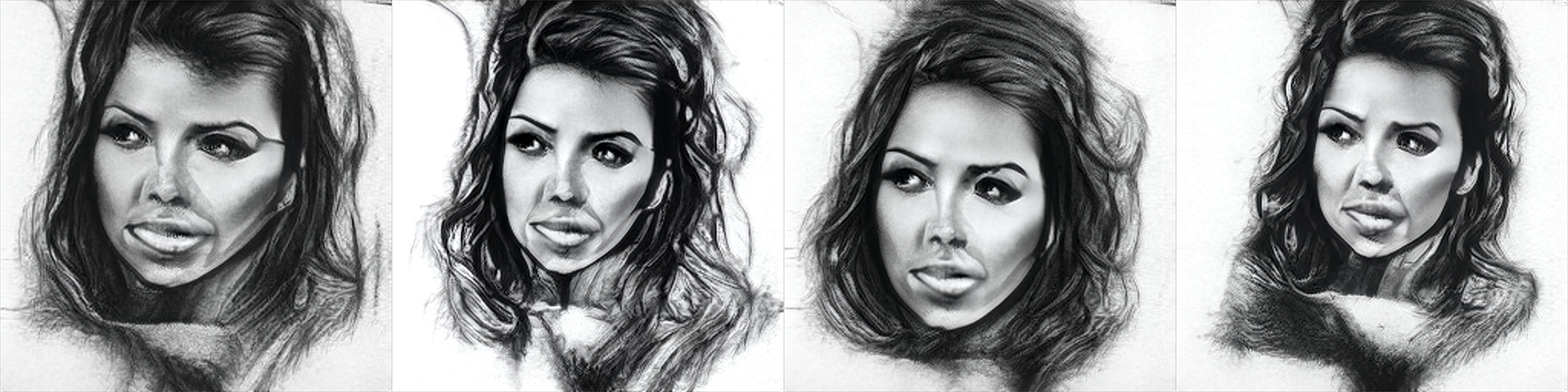}\\
        \scriptsize \makecell{\quad$\psi = 0.10 $\\\\$\overline{CS} = 0.39$}\;&
        \includegraphics[align=c, width=.83\linewidth]{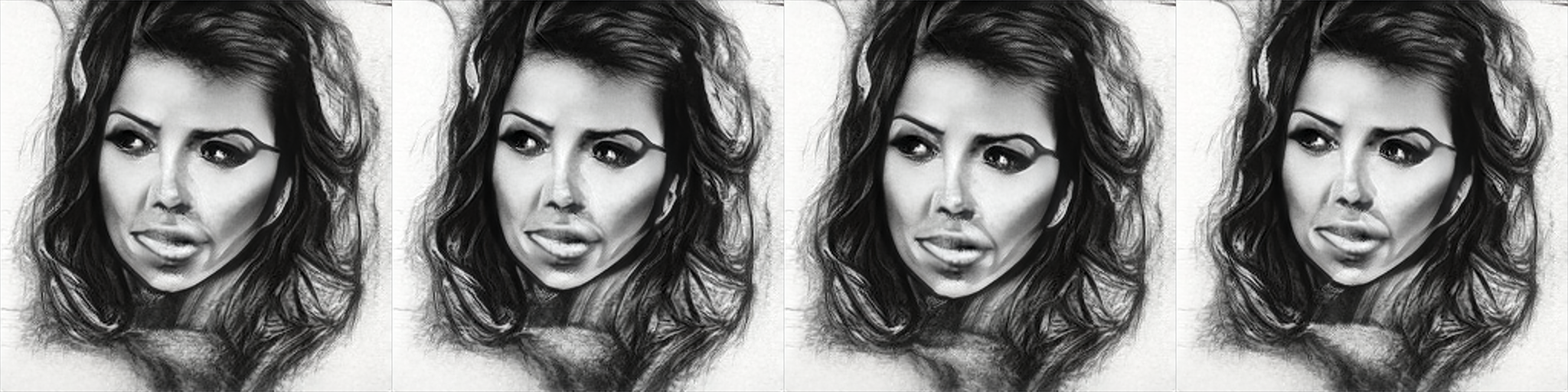}\\
    \end{tabular}        %
\vspace*{-2mm}%
\caption{ 
  \textbf{Truncation.}
  Four samples for the prompt ``a graphite sketch of Eva Longoria'' with different random $\bz$. Increasing truncation (decreasing $\psi$) improves the text alignment according to mean CLIP score per row ($\overline{CS}$) at the cost of lower variation.
}%
\vspace*{-2mm}%
\label{fig:truncationsamples}
\end{figure}
}

\newcommand{\editinterpolations}{
\begin{figure}[t]
\centering
\includegraphics[width=\linewidth]{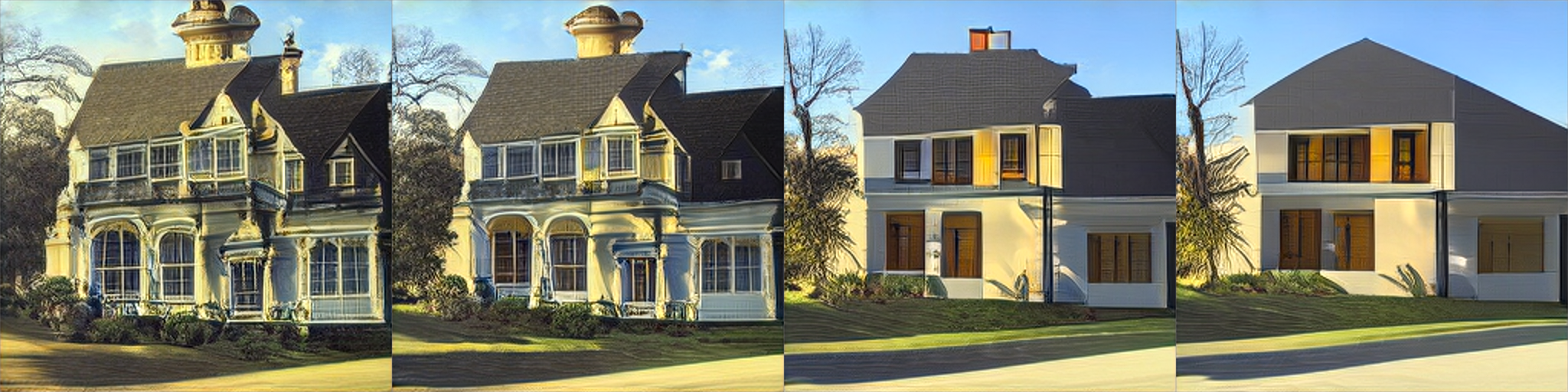} \\
\scriptsize \makecell{``a victorian house'' \quad $\rightarrow$ \quad ``a modern house''}
\includegraphics[width=\linewidth]{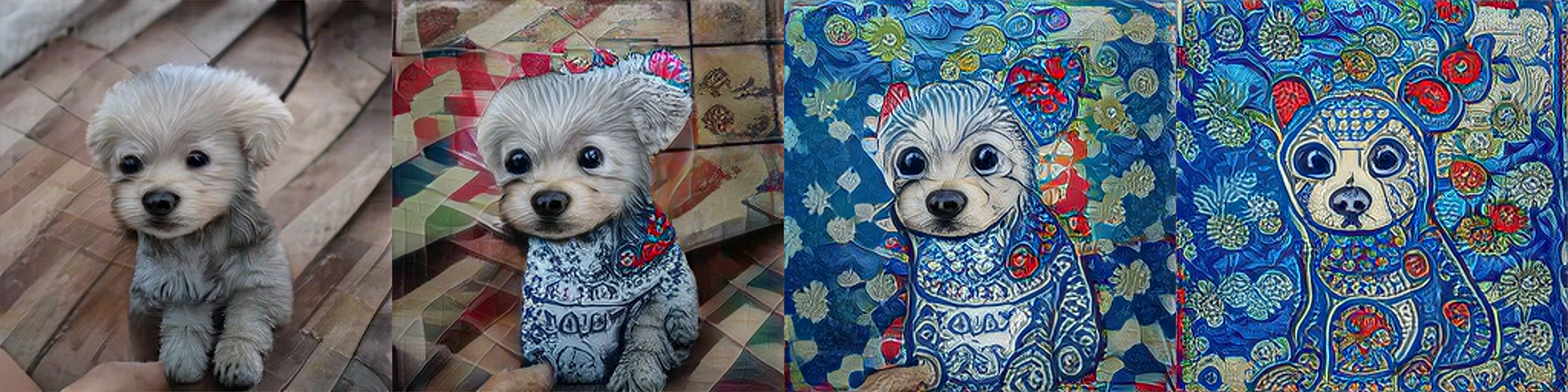} \\
\scriptsize \makecell{``a cute puppy'' \quad $\rightarrow$ \quad ``a cute blue puppy, Madhubani painting''}
\includegraphics[width=\linewidth]{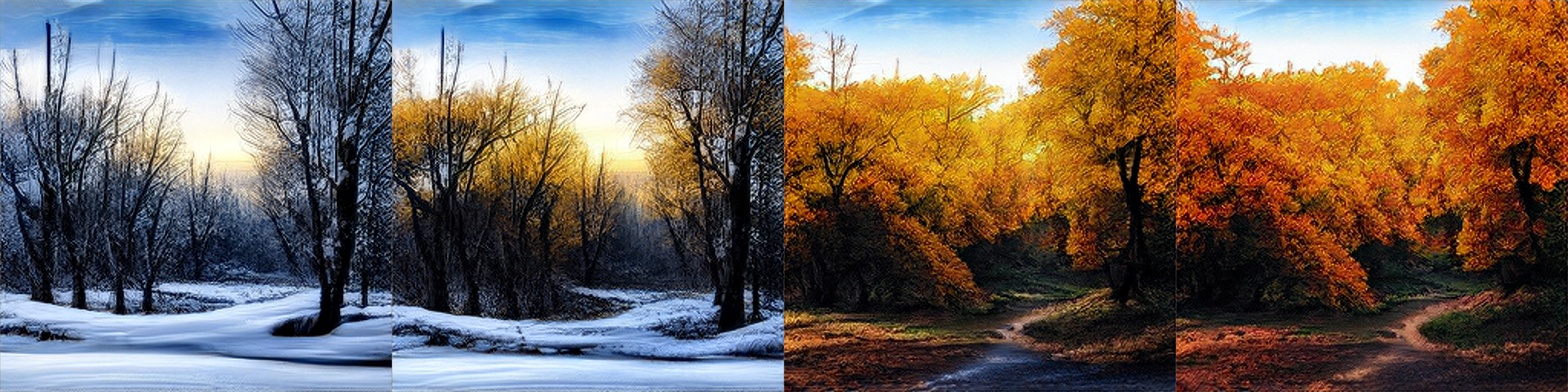} \\
\scriptsize \makecell{``a landscape in winter'' \quad $\rightarrow$ \quad ``a landscape in fall''}%
\vspace*{-2mm}%
\caption{ 
\textbf{Latent manipulation.} Samples (first column) can be manipulated by following semantic directions in latent space.
}
\vspace*{-2mm}%
\label{fig:editinterpolations}
\end{figure}
}

\newcommand{\failures}{
\begin{figure}[t]
    \centering
    \setlength{\tabcolsep}{0.2pt}
    \begin{tabular}{ccc}
        \includegraphics[width=.33\linewidth]{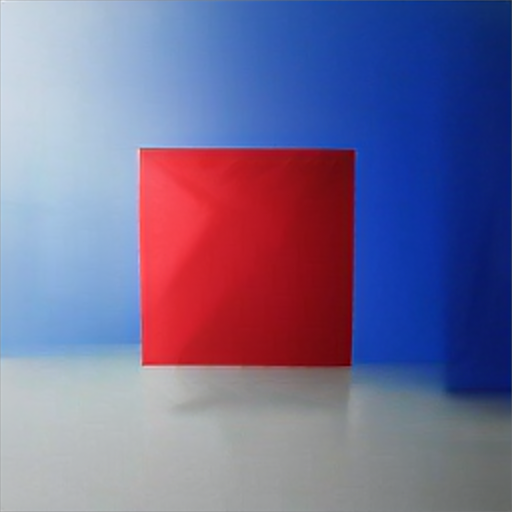} & \includegraphics[width=.33\linewidth]{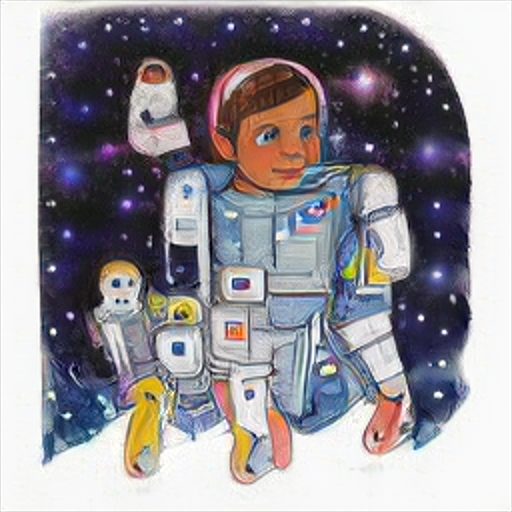} & \includegraphics[width=.33\linewidth]{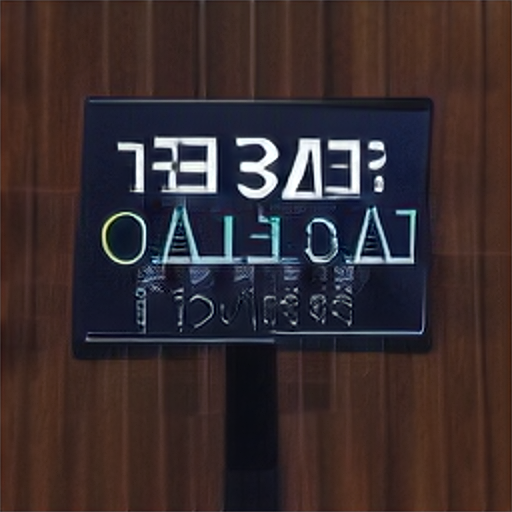} \\
        \scriptsize \makecell{``a red cube on \\a blue cube''} &
        \scriptsize \makecell{``astronaut, child's\\ drawing''} &
        \scriptsize \makecell{``a sign that says \\deep learning''}
    \end{tabular}%
\vspace*{-2mm}%
\caption{ 
  \textbf{Failure cases.}
  StyleGAN-T can struggle to bind attributes to objects,  and to produce coherent text.
}%
\vspace*{-2mm}%
\label{fig:failures}
\end{figure}
}

\newcommand{\truncationsupp}{
\begin{figure}[t]
\centering
\setlength{\tabcolsep}{0.2pt}
\begin{tabular}{ll@{\hspace{1em}}ll}
\tiny \makecell{\quad$\psi = 1.00 $\\\\$\overline{CS} = 0.34$}\;& 
\includegraphics[align=c, width=.414\linewidth]{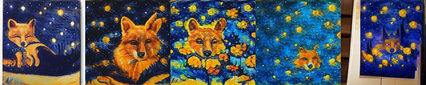}& 
\tiny \makecell{\quad$\psi = 1.00 $\\\\$\overline{CS} = 0.30    $}\;&
\includegraphics[align=c, width=.414\linewidth]{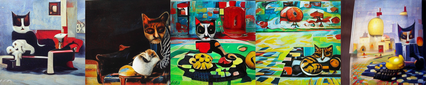}\\
\tiny \makecell{\quad$\psi = 0.60 $\\\\$\overline{CS} = 0.36$}\;& 
\includegraphics[align=c, width=.414\linewidth]{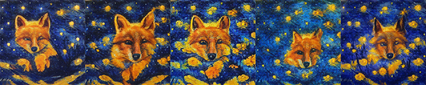}& 
\tiny \makecell{\quad$\psi = 0.60 $\\\\$\overline{CS} = 0.31$}\;&
\includegraphics[align=c, width=0.414\linewidth]{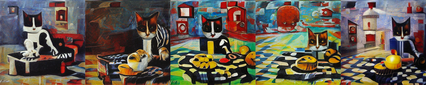}\\
\tiny \makecell{\quad$\psi = 0.10 $\\\\$\overline{CS} = 0.36$}\;& 
\includegraphics[align=c, width=0.414\linewidth]{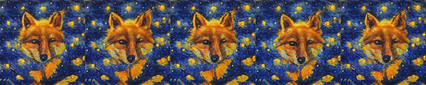}& 
\tiny \makecell{\quad$\psi = 0.10 $\\\\$\overline{CS} = 0.32$}\;&
\includegraphics[align=c, width=0.414\linewidth]{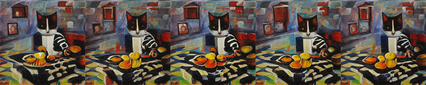}
\vspace{0.5em}\\
\multicolumn{2}{c}{\scriptsize``A painting of a fox in the style of starry night''}
&
\multicolumn{2}{c}{\scriptsize``A surrealist dream-like oil painting by Salvador Dal\'{i} of a cat playing checkers''} \\\\
\tiny \makecell{\quad$\psi = 1.00 $\\\\$\overline{CS} = 0.26$}\;& 
\includegraphics[align=c, width=.414\linewidth]{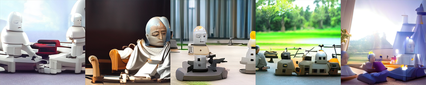}& 
\tiny \makecell{\quad$\psi = 1.00 $\\\\$\overline{CS} = 0.32    $}\;&
\includegraphics[align=c, width=.414\linewidth]{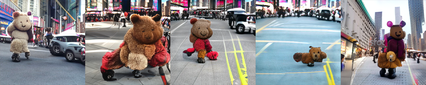}\\
\tiny \makecell{\quad$\psi = 0.60 $\\\\$\overline{CS} = 0.28$}\;& 
\includegraphics[align=c, width=.414\linewidth]{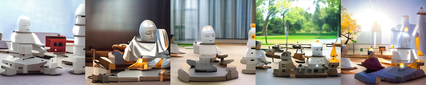}& 
\tiny \makecell{\quad$\psi = 0.60 $\\\\$\overline{CS} = 0.33$}\;&
\includegraphics[align=c, width=0.414\linewidth]{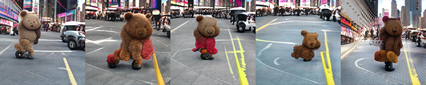}\\
\tiny \makecell{\quad$\psi = 0.10 $\\\\$\overline{CS} = 0.31$}\;& 
\includegraphics[align=c, width=0.414\linewidth]{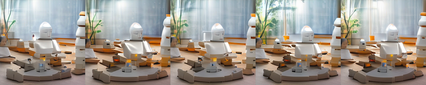}& 
\tiny \makecell{\quad$\psi = 0.10 $\\\\$\overline{CS} = 0.34$}\;&
\includegraphics[align=c, width=0.414\linewidth]{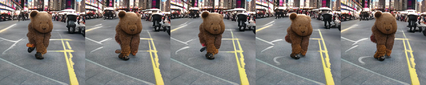}
\vspace{0.5em}\\
\multicolumn{2}{c}{\scriptsize``Robots meditating in a vipassana retreat''} &
\multicolumn{2}{c}{\scriptsize``A teddy bear on a skateboard in times square''}\\\\
\tiny \makecell{\quad$\psi = 1.00 $\\\\$\overline{CS} = 0.28$}\;& 
\includegraphics[align=c, width=0.414\linewidth]{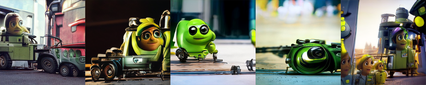}& 
\tiny \makecell{\quad$\psi = 1.00 $\\\\$\overline{CS} = 0.39$}\;&
\includegraphics[align=c, width=0.414\linewidth]{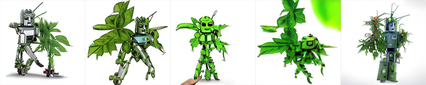}\\
\tiny \makecell{\quad$\psi = 0.60 $\\\\$\overline{CS} = 0.29$}\;& 
\includegraphics[align=c, width=0.414\linewidth]{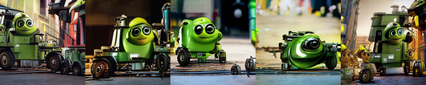}& 
\tiny \makecell{\quad$\psi = 0.60 $\\\\$\overline{CS} = 0.40$}\;&
\includegraphics[align=c, width=0.414\linewidth]{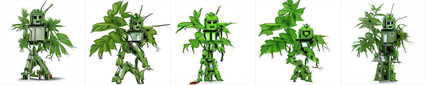}\\
\tiny \makecell{\quad$\psi = 0.10 $\\\\$\overline{CS} = 0.30$}\;& 
\includegraphics[align=c, width=0.414\linewidth]{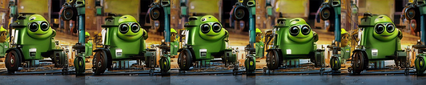}& 
\tiny \makecell{\quad$\psi = 0.10 $\\\\$\overline{CS} = 0.40$}\;&
\includegraphics[align=c, width=0.414\linewidth]{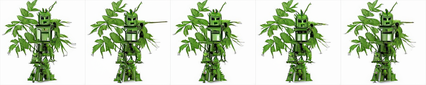}
\vspace{0.5em}\\
\multicolumn{2}{c}{\scriptsize``A still of Kermit The Frog in WALL-E (2008)''} &
\multicolumn{2}{c}{\scriptsize``A transformer robot with legs and arms made out of vegetation and leaves''}
\end{tabular}      
\caption{ 
  \textbf{Additional truncation grids.}
  We show samples for 6 different prompts and 5 different random latents, shared between the prompts. Increasing truncation (decreasing $\psi$), improves the text alignment according to mean CLIP score per row, $\overline{CS}$, at the cost of lower variation.
}
\label{fig:truncationsupp}
\end{figure}
}

\newcommand{\s}{\hphantom{0}}

\newcommand{\ablationtable}{
\begin{table}[t]
\centering
\begin{tabular}{@{}lcc@{}}
\toprule
                      & Zero-shot FID\textsubscript{30k} $\downarrow$ & CLIP score $\uparrow$\\ \midrule
StyleGAN-XL           & 51.88 & \s5.58 \\
New generator         & 45.10 & \s6.02 \\
New discriminator     & 26.77 & \s9.78 \\
$\mathcal{L}_\mathrm{CLIP}$  & \textbf{20.52} & \textbf{11.72} \\ \bottomrule
\end{tabular}
\caption{
\textbf{Architecture ablation.}
Our architectural changes notably improve sample quality and text alignment. Here, we use the lightweight training configuration described in Appendix~\ref{sec:details}. 
}
\label{tab:ablation}
\end{table}
}

\newcommand{\cocotablelowres}{
\begin{table}[t]
\resizebox{\columnwidth}{!}{%
\begin{tabular}{@{}llcc@{}}
\toprule
Model              & Model type     & \multicolumn{1}{l}{Zero-shot FID\textsubscript{30k}} & \multicolumn{1}{l}{Speed [s]} \\ \midrule
Stable Diffusion * & Diffusion      & \s8.40 & --   \\
eDiff-I            & Diffusion      & \s7.60 & 26.0 \\ 
LDM *              & Diffusion      & \s7.59 & --   \\
GLIDE              & Diffusion      & \s7.40 & 10.9 \\
\midrule
LAFITE *           & GAN            & 14.80  & \bf{\hspace*{-2.7mm}$\sim$0.01}  \\
StyleGAN-T         & GAN             & \bf{\s7.30} & 0.06\\ \bottomrule
\end{tabular}
}\vspace{0.5mm}\\%
\makebox[0.65\linewidth]{\scriptsize{* downsampled to 64$\times$64 pixels using Lanczos}}\hfill%
\makebox[0.30\linewidth]{\scriptsize{-- not available}}%
\vspace{-2mm}
\caption{
\textbf{Comparison of FID on MS COCO 64$\boldsymbol{\times}$64.}
Inference speeds are measured on an A100. 
For LAFITE we estimate what its speed would be at a native 64$\times$64 resolution.
}
\vspace*{-2mm}%
\label{tab:cocolowres}
\end{table}
}

\newcommand{\cocotablelargeres}{
\begin{table}[t]
\resizebox{\columnwidth}{!}{%
\begin{tabular}{@{}llcc@{}}
\toprule
Model              & Model type     & \multicolumn{1}{l}{Zero-shot FID$_{\text{30k}}$} & \multicolumn{1}{l}{Speed [s]} \\ \midrule
LDM                & Diffusion      & 12.63  & \s3.7 \\
GLIDE              & Diffusion      & 12.24  & 15.0  \\
DALL·E 2           & Diffusion      & 10.39  & \s--     \\
Stable Diffusion * & Diffusion      & \s8.59 & \s3.7 \\
Imagen            & Diffusion      & \s7.27 & \s9.1 \\
eDiff-I           & Diffusion      & \bf{\s6.95} & 32.0  \\ \midrule
DALL·E            & Autoregressive & 27.50  & \s--     \\
Ernie-ViLG        & Autoregressive & 14.70  & \s--     \\
Make-A-Scene *    & Autoregressive & 11.84  & 25.0  \\
Parti-3B          & Autoregressive & \s8.10 & \s6.4 \\ 
Parti-20B         & Autoregressive & \s7.23 & \s--     \\ \midrule
LAFITE            & GAN            & 26.94  & \bf{0.02}  \\
StyleGAN-T *      & GAN            & 13.90  & 0.10  \\ \bottomrule
\end{tabular}
}\vspace{0.5mm}\\%
\makebox[0.65\linewidth]{\scriptsize{* downsampled to 256$\times$256 pixels using Lanczos}}\hfill%
\makebox[0.30\linewidth]{\scriptsize{-- not available}}%
\vspace{-2mm}
\caption{
\textbf{Comparison of FID on MS COCO 256$\boldsymbol{\times}$256.}
Inference speeds are measured on an A100, except for Imagen and Parti that use a faster TPUv4 accelerator. The Stable Diffusion numbers are from~\cite{balaji2022ediffi,sdinference}; the other numbers are obtained from the respective papers or through correspondence with the authors.
}
\vspace*{-2mm}%
\label{tab:cocohighres}
\end{table}
}

\newcommand{\architecturebase}{
\begin{table}[h]
\centering
\resizebox{0.90\linewidth}{!}{%
\begin{tabular}{lcc}
\hline
 & \textbf{Lightweight} & \textbf{Full} \\ \hline
Generator channel base & 32768 & 65536 \\
Generator channel max & 512 & 2048 \\
Number of residual blocks per generator block & 3 & 4 \\
Generator parameters & 75 million & 1.02 billion \\
Text encoder parameters & 123 million & 123 million \\
Latent (z) dimension & 64 & 64 \\ \hline
Discriminator's feature network & DINO ViT-S/16 & DINO ViT-S/16 \\
Discriminator head's input feature space size & 384 & 384 \\
Discriminator head's feature space size at text conditioning\hspace{2mm}  & 64 & 64 \\ \hline
Dataset size & 12M & ~250M \\
Number of GPUs & 8 & 64 \\
Batch size & 2048 & 2048 \\
Optimizer & Adam & Adam \\
Generator learning rate & 0.002 & 0.002 \\
Generator Adam betas & (0, 0.99) & (0, 0.99) \\
Discriminator learning rate & 0.002 & 0.002 \\
Discriminator Adam betas & (0, 0.99) & (0, 0.99) \\
EMA & 0.9978 & 0.9978 \\
CLIP guidance weight & 0.2 & 0.2 (primary phase), 50 (secondary phase)\\
Progressive growing & No & Yes \\\hline
\end{tabular}
}
\caption{
Generator, discriminator, and training hyperparameters for the two setups used in this paper: Lightweight and Full configuration. 
}
\label{tab:arch_base}
\end{table}
}

\newcommand{\trainingbase}{
\begin{table}[h]
\centering
\resizebox{0.725\linewidth}{!}{%
\begin{tabular}{cc}
\hline
\textbf{Lightweight} & \textbf{Full}\\ \hline
\begin{tabular}[t]{@{}c@{}}
Primary Phase\\ 64x64 for 50 A100 days \hspace{0.1mm} (25 million iterations)\hspace{5mm}
\end{tabular} & 
\begin{tabular}[t]{@{}c@{}}
Primary Phase\\ 
16x16 for 450 A100 days \hspace{0.1mm} (118,000 iterations) \\ %
32x32 for 450 A100 days \hspace{0.1mm} (\s78,000 iterations)\\  %
64x64 for 450 A100 days \hspace{0.1mm} (\s57,000 iterations)\\  %
\\ 
Secondary Phase\\ 
190 A100 days \hspace{0.1mm} (20,000 iterations)\\  %
\\ 
Primary Phase\\ 
128x128 for 96 A100 days \hspace{0.1mm} (10,000 iterations)\\  %
256x256 for 70 A100 days \hspace{0.1mm} (\s6,000 iterations)\\  %
512x512 for 30 A100 days \hspace{0.1mm} (\s3,000 iterations)   %
\end{tabular} \\\hline
\end{tabular}
}
\caption{
Training schedules for the two training configurations used in this paper. The times are listed as the number of days it would have taken on a single NVIDIA A100 GPU. An iteration corresponds to 2048 real and generated examples. 
}
\label{tab:train_base}
\end{table}
}
\begin{document}

\twocolumn[
\icmltitle{\texorpdfstring
{StyleGAN-T: Unlocking the Power of GANs for\\Fast Large-Scale Text-to-Image Synthesis}
{StyleGAN-T: Unlocking the Power of GANs for Fast Large-Scale Text-to-Image Synthesis}}

\icmlsetsymbol{equal}{*}

\begin{icmlauthorlist}
\icmlauthor{Axel Sauer}{utmpi,nv}
\icmlauthor{Tero Karras}{nv}
\icmlauthor{Samuli Laine}{nv}
\icmlauthor{Andreas Geiger}{utmpi}
\icmlauthor{Timo Aila}{nv}
\end{icmlauthorlist}

\icmlaffiliation{utmpi}{University of T{\"u}bingen, T{\"u}bingen AI Center}
\icmlaffiliation{nv}{NVIDIA}

\icmlcorrespondingauthor{Axel Sauer}{a.sauer@uni-tuebingen.de}

\icmlkeywords{Machine Learning, ICML}

\vskip 0.3in
]

\begin{NoHyper}
\printAffiliationsAndNotice{}  %
\end{NoHyper}
\linepenalty=1000

\begin{abstract}
Text-to-image synthesis has recently seen significant progress thanks to large pretrained language models, large-scale training data, and the introduction of scalable model families such as diffusion and autoregressive models.
However, the best-performing models require iterative evaluation to generate a single sample.
In contrast, generative adversarial networks (GANs) only need a single forward pass. They are thus much faster, but they currently remain far behind the state-of-the-art in large-scale text-to-image synthesis.
This paper aims to identify the necessary steps to regain competitiveness.
Our proposed model, \mbox{StyleGAN-T}, addresses the specific requirements of large-scale text-to-image synthesis, such as large capacity, stable training on diverse datasets, strong text alignment, and controllable variation vs.~text alignment tradeoff.
StyleGAN-T significantly improves over previous GANs and outperforms distilled diffusion models\,---\,the previous state-of-the-art in fast text-to-image synthesis\,---\,in terms of sample quality and speed.
\end{abstract}

\section{Introduction}\label{sec:introduction}

In text-to-image synthesis, novel images are generated based on text prompts. 
The state-of-the-art in this task has recently taken dramatic leaps forward thanks to two key ideas. 
First, using a large pretrained language model as an encoder for the prompts makes it possible to condition the synthesis based on general language understanding \cite{ramesh2022hierarchical,saharia2022photorealistic}.
Second, using large-scale training data consisting of hundreds of millions of image-caption pairs \cite{schuhmann2022laion} allows the models to synthesize almost anything imaginable.

\teaserlog

Training datasets continue to increase rapidly in size and coverage.
Consequently, text-to-image models must be scalable to a large capacity to absorb the training data.
Recent successes in large-scale text-to-image generation have been driven by diffusion models (DM)~\cite{ramesh2022hierarchical, saharia2022photorealistic, rombach2022high} and autoregressive models (ARM)~\cite{zhang2021ernie, yu2022scaling,gafni2022make} that seem to have this property built in, along with the ability to deal with highly multi-modal data.

Interestingly, generative adversarial networks (GAN) \cite{Goodfellow2014}\,---\,the dominant family of generative models in smaller and less diverse datasets\,---\,have not been particularly successful in this task \cite{zhou2022towards}.
Our goal is to show that they can regain competitiveness.

The primary benefits offered by GANs are inference speed and control of the synthesized result via latent space manipulations. %
StyleGAN \cite{karras2019style,karras2020analyzing,karras2021alias} in particular has a thoroughly studied latent space, which allows principled control of generated images \cite{Bermano2022,Harkonen2020,shen2020interpreting,Abdal2021,Kafri2021}.
While there has been notable progress in speeding up DMs~\cite{salimans2022progressive,karras2022elucidating,lu2022dpm}, they are still far behind GANs that require only a single forward pass. %

We draw motivation from the observation that GANs lagged similarly behind diffusion models in ImageNet \cite{deng2009imagenet,dhariwal2021diffusion} synthesis until the discriminator architecture was redesigned in StyleGAN-XL \cite{sauer2021projected,sauer2022stylegan}, which allowed GANs to close the gap. 
In Section~\ref{sec:arch}, we start from StyleGAN-XL and revisit the generator and discriminator architectures, considering the requirements specific to the large-scale text-to-image task: large capacity, extremely diverse datasets, strong text alignment, and controllable variation~vs.~text alignment tradeoff.

We have a fixed training budget of 4 weeks on 64 NVIDIA A100s available for training our final model at scale. 
This constraint forces us to set priorities because the budget is likely insufficient for state-of-the-art, high-resolution results \cite{sdtrainingtime}.
While the ability of GANs to scale to high resolutions is well known \cite{wang2018esrgan,karras2020analyzing}, successful scaling to the large-scale text-to-image task remains undocumented.
We thus focus primarily on solving this task in lower resolutions, dedicating only a limited budget to the super-resolution stages. 

Our StyleGAN-T achieves a better zero-shot MS COCO FID \cite{lin2014microsoft, heusel2017gans} than current state-of-the-art diffusion models at a resolution of 64$\times$64.
At 256$\times$256, StyleGAN-T halves the zero-shot FID previously achieved by a GAN but continues to trail SOTA diffusion models.
The key benefits of StyleGAN-T include its fast inference speed and smooth latent space interpolation in the context of text-to-image synthesis, illustrated in \figref{fig:teaser} and \figref{fig:bigsamples}, respectively.
\ifgithub
We will make our implementation available at \url{https://github.com/autonomousvision/stylegan-t}
\else
We will make our implementation publicly available.
\fi

\bigsamples

\section{StyleGAN-XL}\label{sec:background}

Our architecture design is based on StyleGAN-XL~\cite{sauer2022stylegan} that\,---\,similar to the original StyleGAN~\cite{karras2019style}\,---\,first processes the normally distributed input latent code~$\bz$ by a mapping network to produce an intermediate latent code~$\bw$.
This intermediate latent is then used to modulate the convolution layers in a synthesis network using the weight demodulation technique introduced in StyleGAN2~\cite{karras2020analyzing}.
The synthesis network of StyleGAN-XL uses the alias-free primitive operations of StyleGAN3~\cite{karras2021alias} to achieve translation equivariance, i.e., to enforce the synthesis network to have no preferred positions for the generated features.

StyleGAN-XL has a unique discriminator design where multiple discriminator heads operate on feature projections~\cite{sauer2021projected} from two frozen, pretrained feature extraction networks: \mbox{DeiT-M}~\cite{touvron2021deit} and EfficientNet~\cite{tan2019efficientnet}. 
Their outputs are fed through randomized cross-channel and cross-scale mixing modules.
This results in two feature pyramids with four resolution levels each that are then processed by eight discriminator heads.
An additional pretrained classifier network is used to provide guidance during training.

The synthesis network of StyleGAN-XL is trained progressively, increasing the output resolution over time by introducing new synthesis layers once the current resolution stops improving.
In contrast to a previous progressive growing approach~\cite{karras2017progressive}, the discriminator structure does not change during training.
Instead, the early low-resolution images are upsampled as necessary to suit the discriminator.
In addition, the already trained synthesis layers are frozen as further layers are added.

For class-conditional synthesis, StyleGAN-XL concatenates an embedding of a one-hot class label to $\bz$ %
and uses a projection discriminator \cite{miyato2018cgans}.

\section{StyleGAN-T}\label{sec:method}
\label{sec:arch}

\ablationtable
\system

We choose StyleGAN-XL as our baseline architecture because of its strong performance in class-conditional ImageNet synthesis \cite{sauer2022stylegan}.
In this section, we modify this baseline piece by piece, focusing on the generator (Section~\ref{sec:generator}), discriminator (Section~\ref{sec:discriminator}), and variation vs.~text alignment tradeoff mechanisms~(Section \ref{sec:guidance}) in turn.

Throughout the redesign process, we measure the effect of our changes using zero-shot MS COCO. For practical reasons, the tests use a limited compute budget, smaller models, and a smaller dataset than the large-scale experiments in Section~\ref{sec:experiments}; see Appendix~\ref{sec:details} for details.
We quantify sample quality using FID \cite{heusel2017gans} and text alignment using CLIP score \cite{hessel2021clipscore}.
Following prior art \cite{balaji2022ediffi}, we compute the CLIP score using a \mbox{ViT-g-14} model trained on \mbox{LAION-2B}~\mbox{\cite{schuhmann2022laion}}.

To change the class conditioning to text conditioning in our baseline model, 
we embed the text prompts using a pretrained CLIP \mbox{ViT-L/14} text encoder \cite{radford2021learning} and use them in place of the class embedding.
Accordingly, we also remove the training-time classifier guidance.
This simple conditioning mechanism matches the early text-to-image models \cite{reed16neurips,reed16icml}.
As shown in \tabref{tab:ablation}, this baseline reaches a zero-shot FID of 51.88 and CLIP score of 5.58 in our lightweight training configuration.
Note that we use a different CLIP model for conditioning the generator and for computing the CLIP score, which reduces the risk of artificially inflating the results.

\subsection{Redesigning the Generator}
\label{sec:generator}

StyleGAN-XL uses StyleGAN3 layers to achieve translational equivariance. 
While equivariance can be desirable for various applications, we do not expect it to be necessary for text-to-image synthesis because none of the successful DM/ARM-based methods are equivariant.
Additionally, the equivariance constraint adds computational cost and poses certain limitations to the training data that large-scale image datasets typically violate~\cite{karras2021alias}.

For these reasons, we drop the equivariance and switch to StyleGAN2 backbone for the synthesis layers, including output skip connections and spatial noise inputs that facilitate stochastic variation of low-level details.
The high-level architecture of our generator after these changes is shown in \figref{fig:system}a. 
We additionally propose two changes to the details of the generator architecture (\figref{fig:system}b).

\boldparagraph{Residual convolutions.} As we aim to increase the model capacity significantly, the generator must be able to scale in both width and depth. However, in the basic configuration, a significant increase in the generator's depth leads to an early mode collapse in training.
An important building block in modern CNN architectures~\cite{liu2022convnet, dhariwal2021diffusion} is an easily optimizable residual block that normalizes the input and scales the output.
Following these insights, we make half the convolution layers residual and wrap them by GroupNorm~\cite{wu2018group} for normalization and Layer Scale~\cite{touvron2021going} for scaling their contribution. %
A layer scale of a low initial value of $10^{-5}$ allows gradually fading in the convolution layer's contribution, 
  stabilizing the early training iterations significantly.
This design allows us to increase the total number of layers considerably\,---\,by approximately 2.3$\times$ in the lightweight configuration and 4.5$\times$ in the final model. For fairness, we match the parameter count of the StyleGAN-XL baseline.

\boldparagraph{Stronger conditioning.} 
The text-to-image setting is challenging because the factors of variation can vastly differ per prompt. Consider the prompts ``a close-up of a face''  and  ``a beautiful landscape.'' The first prompt should generate faces with varying eye color, skin color, and proportions, whereas the second should produce landscapes from different areas, seasons, and daytime.
In a style-based architecture, all of this variation has to be implemented by the per-layer styles. Thus the text conditioning may need to affect the styles much more strongly than was necessary for simpler settings.

In early tests, we observed a clear tendency of the input latent $\bz$ to dominate over the text embedding $\bc_\mathrm{text}$ in our baseline architecture, leading to poor text alignment. 
To remedy this, we introduce two changes that aim to amplify the role of $\bc_\mathrm{text}$.
First, we let the text embeddings bypass the mapping network, following the observations by \citet{Harkonen2022}. 
A similar design was also used in \textsc{Lafite} \cite{zhou2022towards}, assuming that the CLIP text encoder defines an appropriate intermediate latent space for the text conditioning.
We thus concatenate $\bc_\mathrm{text}$ directly to $\bw$ and use a set of affine transforms to produce per-layer styles $\tilde{\bs}$.
Second, instead of using the resulting $\tilde{\bs}$ to modulate the convolutions as-is, we further split it into three vectors of equal dimension $\tilde{\bs}_{1,2,3}$ and compute the final style vector as
\begin{equation}
    \bs = \tilde{\bs}_1 \odot \tilde{\bs}_2 + \tilde{\bs}_3 \text{.}
\label{eq:poly}
\end{equation}
The crux of this operation is the element-wise multiplication~$\odot$ that effectively turns the affine transform into a 2\textsuperscript{nd} order polynomial network \cite{Chrysos2020,Chrysos2021}, increasing its expressive power.
The stacked MLP-based conditioning layers in DF-GAN \cite{tao2022dfgan} implicitly include similar 2\textsuperscript{nd} order terms.

Together, our changes to the generator improve FID and CLIP score by $\sim$10\%, as shown in \tabref{tab:ablation}.

\subsection{Redesigning the Discriminator}
\label{sec:discriminator}
We redesign the discriminator from scratch but retain StyleGAN-XL's key ideas of relying on a frozen, pretrained feature network and using multiple discriminator heads. %

\boldparagraph{Feature network.} 
For the feature network, we choose a \mbox{ViT-S}~\cite{dosovitskiy2020image} trained with the self-supervised DINO objective~\cite{caron2021emerging}. The network is lightweight, fast to evaluate, and encodes semantic information at high spatial resolution~\cite{amir2021deep}.
An additional benefit of using a self-supervised feature network is that it circumvents the concern of potentially compromising FID~\cite{kynkaanniemi2022role}.

\boldparagraph{Architecture.}
Our discriminator architecture is shown in \figref{fig:system}c. 
ViTs are isotropic, i.e., the representation size (tokens $\times$ channels) and receptive field (global) are the same throughout the network.
This isotropy allows us to use the same architecture for all discriminator heads, which we space equally between the transformer layers. 
Multiple heads are known to be beneficial \cite{sauer2021projected}, and we use five heads in our design.

Our discriminator heads are minimalistic, as detailed in \figref{fig:system}c, bottom.
The residual convolution's kernel width controls the head's receptive field in the token sequence. 
We found that 1D convolutions applied on the sequence of tokens performed just as well as 2D convolutions applied on spatially reshaped tokens, 
  indicating that the discrimination task does not benefit from whatever 2D structure remains in the tokens.
We evaluate a hinge loss~\cite{lim2017geometric} independently for each token in every head.

\citet{sauer2021projected} use synchronous BatchNorm~\cite{ioffe2015batch}
to provide batch statistics to the discriminator. %
BatchNorm is problematic when scaling to a multi-node setup, as it requires communication between nodes and GPUs.
We use a variant that computes batch statistics on small virtual batches \cite{hoffer2017train}. The batch statistics are not synchronized between devices but are calculated per local minibatch. 
Furthermore, we do not use running statistics, and thus no additional communication overhead between GPUs is introduced.

\boldparagraph{Augmentation.}
We apply differentiable data augmentation \cite{zhao2020differentiable} with default parameters before the feature network in the discriminator.
We use random crops when training at a resolution larger than 224$\times$224 pixels (\mbox{ViT-S} training resolution).

As shown in Table~\ref{tab:ablation}, these changes significantly improve FID and CLIP score by further $\sim$40\%. 
This considerable improvement indicates that a well-designed discriminator is critical when dealing with highly diverse datasets.
Compared to the StyleGAN-XL discriminator, our simplified redesign is $\sim$2.5$\times$ faster, leading to $\sim$1.5$\times$ faster training.

\subsection{Variation vs.~Text Alignment Tradeoffs}
\label{sec:guidance}

Guidance \cite{dhariwal2021diffusion,ho2022classifier} is an essential component of current text-to-image diffusion models. It trades variation for perceived image quality in a principled way, preferring images that are strongly aligned with the text conditioning.
In practice, guidance drastically improves the results; thus, we want to approximate its behavior in the context of GANs.

\boldparagraph{Guiding the generator.}
StyleGAN-XL uses a pretrained ImageNet classifier to provide additional gradients during training, guiding the generator toward images that are easy to classify.
This method improves results significantly. 
In the context of text-to-image, ``classification'' involves captioning the images. Thus, a natural extension of this approach is to use a CLIP image encoder instead of a classifier.
Following \citet{crowson2022vqgan}, at each generator update, we pass the generated image through the CLIP image encoder to obtain caption $\bc_\mathrm{image}$, and minimize the squared spherical distance to the normalized text embedding $\bc_\mathrm{text}$:
\begin{equation}
\mathcal{L}_\mathrm{CLIP} = \arccos^2 ( \bc_\mathrm{image} \cdot \bc_\mathrm{text} ) %
\label{eq:guidance}
\end{equation}
This additional loss term guides the generated distribution towards images that are captioned similarly to the input text encoding $\bc_\mathrm{text}$. Its effect is thus similar to the guidance in diffusion models.
\figref{fig:system}d illustrates our approach.

CLIP has been used in prior work to guide a pretrained generator during synthesis~\cite{nichol2021glide, crowson2022vqgan, liu2021fusedream}. 
In contrast, we use it as a part of the loss function during training.
It is important to note that overly strong CLIP guidance during training impairs FID,
  as it limits the distribution diversity and ultimately starts introducing image artifacts.
Therefore, the weight of $\mathcal{L}_\mathrm{CLIP}$ in the overall loss needs to strike a balance between image quality, text conditioning, and
  distribution diversity; we set it to 0.2.
We further observed that guidance is helpful only up to 64$\times$64 pixel resolution.
At higher resolutions, we apply $\mathcal{L}_\mathrm{CLIP}$ to random 64$\times$64 pixel crops.

As shown in \tabref{tab:ablation}, CLIP guidance improves FID and CLIP scores by further $\sim$20\%.

\boldparagraph{Guiding the text encoder.}
Interestingly, the earlier methods listed above that use a pretrained generator did not report encountering low-level image artifacts.
We hypothesize that the frozen generator acts as a prior that suppresses them.
We build on this insight to further improve the text alignment. 
In our primary training phase, the generator is trainable and the text encoder is frozen.
We then introduce a secondary phase, where the generator is frozen and the text encoder becomes trainable instead.
We only train the text encoder as far as the generator conditioning is concerned; the discriminator and the guidance term (\shorteqnref{eq:guidance}) still receive $\bc_\mathrm{text}$ from the original frozen encoder.
This secondary phase allows a very high CLIP guidance weight of 50 without introducing artifacts and significantly improves text alignment without compromising FID (\secref{sec:text_alignment}). Compared to the primary phase, the secondary phase can be much shorter. After convergence, we continue with the primary phase.

\boldparagraph{Explicit truncation.}
Typically variation has been traded to higher fidelity in GANs using the truncation trick \cite{marchesi2017truncation,brock2018large,karras2019style},
  where a sampled latent $\bw$ is interpolated towards its mean with respect to the given conditioning input.
This way, truncation pushes $\bw$ to a higher-density region where the model performs better.
In our implementation, $\bw = [f(\bz), ~\bc_\mathrm{text}]$, where $f(\cdot)$ denotes the mapping network, so the per-prompt mean is given by $\tilde{\bw} = \mathbb{E}_\bz[\bw] = [\tilde{\mathbf{f}}, ~\bc_\mathrm{text}]$, where $\tilde{\mathbf{f}} = \mathbb{E}_\bz[f(\bz)]$.
We thus implement truncation by tracking $\tilde{\mathbf{f}}$ during training and interpolating between $\tilde{\bw}$ and $\bw$ according to scaling parameter $\psi \in [0,1]$ at inference time.

We illustrate the impact of truncation in \figref{fig:truncationsamples}. 
In practice, we rely on the combination of CLIP guidance and truncation.
Guidance improves the model's overall text alignment, and truncation can further boost quality and alignment for a given sample, trading away some variation.
\truncationsamples

    \section{Experiments}
\label{sec:experiments}

Using the final configuration developed in \secref{sec:arch}, we scale the model size, dataset, and training time. Our final model consists of $\sim$1 billion parameters; we did not observe any instabilities when increasing the model size.
We train on a union of several datasets 
amounting to 250M text-image pairs in total.
We use progressive growing similar to StyleGAN-XL, except that all layers remain trainable.
The hyperparameters and dataset details are listed in Appendix~\ref{sec:details}.

The total training time was four weeks on 64 A100 GPUs using a batch size of 2048. %
We first trained the primary phase for 3 weeks (resolutions up to 64$\times$64), then the secondary phase for 2 days (text embedding), and finally the primary phase again for 5 days (resolutions up to 512$\times$512).
For comparison, our total compute budget is about a quarter of Stable Diffusion's~\mbox{\cite{sdtrainingtime}}.

\subsection{Quantitative Comparison to State-of-the-Art}
\cocotablelowres
\cocotablelargeres
We use zero-shot MS COCO to compare the performance of our model to the state-of-the-art quantitatively at 64$\times$64 pixel output resolution in \tabref{tab:cocolowres} and 256$\times$256 in \tabref{tab:cocohighres}.
At low resolution, StyleGAN-T outperforms all other approaches in terms of output quality, while being very fast to evaluate. In this test we use the model before the final training phase, i.e., one that produces 64$\times$64 images natively.
At high resolution, StyleGAN-T still significantly outperforms LAFITE but lags behind DMs and ARMs in terms of FID.

These results lead us to two conclusions. 
First, GANs can match or even beat current DMs in large-scale text-to-image synthesis at low resolution. 
Second, a powerful superresolution model is crucial. While FID slightly decreases in \mbox{eDiff-I} when moving from 64$\times$64 to 256$\times$256 (7.60$\rightarrow$6.95), it currently almost doubles in StyleGAN-T. 
Therefore, it is evident that StyleGAN-T's superresolution stage is underperforming, causing a gap to the current state-of-the-art high-resolution results. 
Whether this gap can be bridged simply with additional capacity or longer training is an open question. %

\subsection{Evaluating Variation vs.~Text Alignment}
\label{sec:text_alignment}
We report FID--CLIP score curves in \figref{fig:truncation}. We compare StyleGAN-T to a strong DM baseline (CLIP-conditioned variant of eDiff-I) and a fast, distilled DM baseline (SD-distilled)~\cite{meng2022distillation}.

Using Truncation, StyleGAN-T can push the CLIP score to 0.305, successfully improving text alignment. StyleGAN-T outperforms SD-distilled in both FID and CLIP scores yet remains behind eDiff-I. 
Regarding speed, eDiff-I requires 32.0 seconds to generate a sample.
SD-distilled is significantly faster and only needs 0.6 seconds at its best performance at eight sampling steps.
StyleGAN-T beats both baselines, generating a sample in 0.1 seconds.

\truncation

To isolate the impact of text encoder training, we evaluate FID--CLIP score curves in~\figref{fig:tetraining}. For this experiment, we utilize the same generator network and only swap the text encoder. As the generator has been frozen in the secondary phase, it can handle both the original and fine-tuned CLIP text embeddings as evidenced by their equal performance measured by FID.
Fine-tuning the %
text encoder significantly improves the CLIP score without compromising FID.
\tetraining

\subsection{Qualitative Results}

\figref{fig:bigsamples} shows example images produced by StyleGAN-T, along with interpolations between them. The accompanying video shows this in animation and compares it to diffusion models, demonstrating that the interpolation properties of GANs continue to be considerably smoother. %

Interpolating between different text prompts is straightforward. For an image generated by an intermediate latent $\bw_0 = [f(\bz), ~\bc_{\mathrm{text0}}]$, we substitute the text condition $\bc_{\mathrm{text0}}$ with a new text condition $\bc_{\mathrm{text1}}$. We then interpolate $\bw_0$ towards the new latent $\bw_1 = [f(\bz), ~\bc_{\mathrm{text1}}]$ as shown in \figref{fig:editinterpolations}. This approach is similar to DALL$\cdot$E~2's text diff operation that interpolates between CLIP embeddings.
Previous work for manipulating GAN-generated images \cite{Patashnik_2021} typically discovers these latent directions via a training process that needs to be repeated per prompt and is, therefore, expensive.
Meaningful latent directions are a built-in property of our model, and no extra training is needed.
\editinterpolations

By appending different styles to a prompt, StyleGAN-T can generate a wide variety of styles as shown in \figref{fig:stylesamples}. Subjects tend to be aligned for a fixed latent $\bz$, which we showcase in the accompanying video.
\stylesamples

\section{Limitations and Future Work}
\label{sec:discussion}

Similarly to DALL$\cdot$E~2 that also uses CLIP as the underlying language model, StyleGAN-T sometimes struggles in terms of binding attributes to objects as well as producing coherent text in images (\figref{fig:failures}). Using a larger language model would likely resolve this issue at the cost of slower runtime~\cite{saharia2022photorealistic,balaji2022ediffi}. 
\failures

Guidance via CLIP loss is vital for good text alignment, but high guidance strength results in image artifacts. A possible solution could be to retrain CLIP on higher-resolution data 
that does not suffer from aliasing or other image quality issues. 
In this context, the conditioning mechanism in the discriminator may also be worth revisiting.

Truncation improves text alignment but differs from guidance in diffusion models in two important ways. 
While truncation is always towards a single mode, guidance can at least theoretically be arbitrarily multi-modal. Also, truncation sharpens the distribution before the synthesis network, which can reshape the distribution in arbitrary ways, thus, possibly undoing any prior sharpening.
Therefore, alternative methods to truncation might further improve the results.

Improved super-resolution stages (i.e., high-resolution layers) through higher capacity and longer training are an obvious avenue for future work.

Methods for ``personalizing'' %
diffusion models have become popular~\cite{ruiz2022dreambooth, gal2022image}. They finetune a pretrained model to associate a unique identifier with a given subject, allowing it to synthesize novel images of the same subject in novel contexts.
Such approaches could be similarly applied to GANs.

\clearpage
\ifacknowledgments\section*{Acknowledgements}
We would like to thank Tim Brooks, Miika Aittala, and Jaakko Lehtinen for helpful discussions; Yogesh Balaji and Seungjun Nah for computing additional metrics for eDiff-I; Tuomas Kynk\"{a}\"{a}nniemi for extensive feedback on an earlier draft; Tero Kuosmanen, Samuel Klenberg, and Janne Hellsten for maintaining the compute infrastructure; and David Luebke and Vanessa Sauer for their general support. \fi
\bibliography{bibliography_mid,bibliography_icml}
\bibliographystyle{icml2021}

\clearpage
\appendix
\onecolumn

\section{Configuration Details}
\label{sec:details}

\tabref{tab:arch_base} lists the training and network architecture hyperparameters for our two configurations: lightweight (used for ablations) and the full configuration (used for main results).
\tabref{tab:train_base} details the training schedules.

\boldparagraph{Lightweight training configuration.} 
We train using the CC12M dataset~\cite{changpinyo2021cc12m} at 64$\times$64 resolution, without
using progressive growing.

\boldparagraph{Full training configuration.}
We train using a union of several datasets: CC12m~\cite{changpinyo2021cc12m}, CC~\cite{sharma2018conceptual}, YFCC100m (filtered)~\cite{thomee2016yfcc100m, singh2022flava}, Redcaps~\cite{desai2021redcaps}, \mbox{LAION-aesthetic-6+}~\cite{schuhmann2022laion}. This amounts to a total of 250M text-image pairs.
We use progressive growing similar to StyleGAN-XL, except that all layers remain trainable.
The vast majority of the training budget is spent on resolutions up to 64$\times$64.

\section{Truncation Grids}

\figref{fig:truncationsupp} shows additional examples of truncation. 
\vspace{0.5\baselineskip}

\architecturebase
\trainingbase
\truncationsupp

\end{document}